%% file: occ_icip.tex
\documentclass{article}
\usepackage{spconf,amsmath,graphicx}
\usepackage{times}
\usepackage{epsfig}
\usepackage{graphicx}
\usepackage{amsmath}
\usepackage{amssymb}

\usepackage{algorithmic}
\usepackage{algorithm}
\usepackage{subcaption} 
\usepackage{xcolor}
\usepackage{booktabs} 
\usepackage{multirow}
\usepackage{balance}
\usepackage{float}
\floatstyle{plaintop}
\restylefloat{table}

\usepackage[pagebackref=true,breaklinks=true,letterpaper=true,colorlinks,bookmarks=false]{hyperref}

\title{Context-Aware Automatic Occlusion Removal}
\name{Kumara Kahatapitiya \qquad Dumindu Tissera \qquad Ranga Rodrigo \vspace*{-2mm}}
\address{Department of Electronic and Telecommunication Engineering\\ University of Moratuwa, Sri Lanka  \vspace*{-2mm}}

\begin{document}

\maketitle

\input{abstract}

\begin{keywords}
	Deep Learning, Context-Awareness, Occlusion Removal \vspace{-4mm}
\end{keywords}

\section{Introduction}
\input{introduction}

\section{Related Work}
\label{se:relatedwork}
\input{relatedwork}

\section{System Architecture}
\label{se:systemarchitecture}
\input{systemarchitecture}

\section{Results}
\label{se:evaluation}
\input{evaluation}

\section{Conclusion}
\label{se:conclusion}
\input{conclusion}

\balance{}
\bibliographystyle{IEEEbib}
\bibliography{bibliography}

\end{document}

%% file: abstract.tex
\begin{abstract}
Occlusion removal is an interesting application of image enhancement, for which, existing work suggests manually-annotated or domain-specific occlusion removal. No work tries to address automatic occlusion detection and removal as a context-aware generic problem. In this paper, we present a novel methodology to identify objects that do not relate to the image context as occlusions and remove them, reconstructing the space occupied coherently. The proposed system detects occlusions by considering the relation between foreground and background object classes represented as vector embeddings, and removes them through inpainting. We test our system on COCO-Stuff dataset and conduct a user study to establish a baseline in context-aware automatic occlusion removal.\vspace{0mm}
\end{abstract} 

%% file: introduction.tex
\vspace{-2mm}Automatically removing unrelated foreground objects from an image to ``clean'' it, is an intriguing and still-open area of research. Images tend to loose the naturalness or the visually-pleasing nature due to the presence of such objects. For instance, a scene of a landscape may contain an artificial object in the foreground, or a picturesque attraction may be crowded, making it impossible to capture its untainted beauty. Such complications give rise to the requirement of automatic detection and removal of undesired or unrelated objects in images. We refer to such objects as \textit{occlusions}. In a general sense, this task  is challenging since the occlusions are dependent on the image context. For instance, an occlusion in one image may be perfectly natural in another.

The process of acquiring an occlusion-free image involves two subtasks: identifying unrelated things, and removing them coherently. Image context defines which objects are unrelated. Yet, due to high complexity of natural images, proper interpretation of the image context is difficult \cite{shotton2009textonboost,Yu_2018_CVPR}. Being unrelated is usually subjective in human perception. However, in a computer vision system, this should be captured objectively. Object detection coupled with scene understanding and pixel generation can potentially address these subtasks.

Since Convolutional Neural Networks (CNNs) enable  rich feature extraction that capture complex contextual information \cite{he2016deep,Dolhansky_2018_CVPR,Liu_2018_CVPR}, and Generative Adversarial Networks (GANs) provide considerable improvements in pixel generation tasks \cite{isola2017image,Yu_2018_CVPR}, we are motivated in adapting neural networks to manipulate contextual information. However, we further require a methodology of identifying unrelated objects based on the image context. It is difficult to train an end-to-end system to interpret the image context and remove occlusions due to a public dataset on \textit{relation} --- how related each object is to a certain image context --- being unavailable. Thus, there is no complete system available which makes intelligent decisions on object relations and removes occlusions, producing a visually-pleasing and occlusion-free image.
\begin{figure}[t]
	\centering
	\begin{subfigure}[t]{0.22\textwidth}
		\centering
		\includegraphics[width=0.994\textwidth]{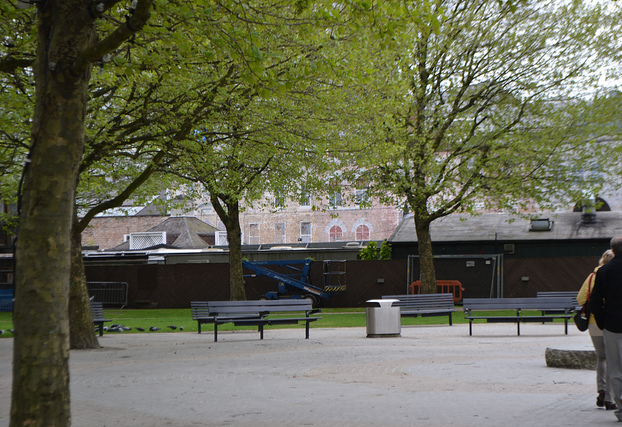}
		\label{Fig:intro_in}
	\end{subfigure}%
	~
	\begin{subfigure}[t]{0.22\textwidth}
		\centering
		\includegraphics[width=1\textwidth]{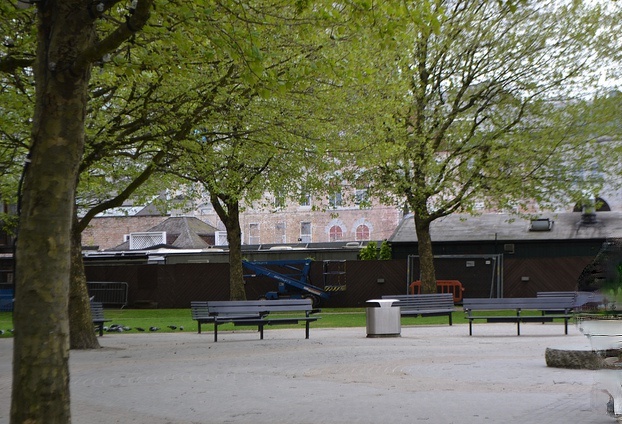}
		\label{Fig:intro_out}
	\end{subfigure}%
	\vspace{-6mm}
	\caption{An example result of the context-aware occlusion removal using our system, without any post-processing. Left: input image, Right: output of the system. Notice that the people and the handbag have been removed as occlusions.}\vspace{-5mm}
	\label{Fig:intro}
\end{figure}

In this paper, we propose a novel approach for occlusion removal using an architecture which fuses both image and language processing. 
First, we process the input image to extract background and foreground object classes separately with pixel-wise annotations of foreground objects. Then, our language model intuitively decides the relation of each foreground
object to the image context; hence, it identifies occlusions. Finally, we mask the pixels of occlusions and feed the image into an inpainting model which produces an occlusion-free image. This task separation allows us to tackle the issue of the lack of object-background relationships in datasets, since our system relies on semantic segmentations and image captions for training.
An example result of our system is shown in Fig.\ref{Fig:intro}.

Contributions of this paper are as follows:\vspace{-2mm}
\begin{itemize}
	\item We propose a novel system to automatically detect occlusions based on image context and remove them, producing a visually-pleasing occlusion-free image.\vspace{-2mm}
	\item We present the first approach which makes intelligent decisions on the relation of objects to image context, based on foreground- and background-object detection.\vspace{-2mm}
\end{itemize} 

%% file: relatedwork.tex
\vspace{-4mm}
Previous work related to context-aware occlusion removal includes pre-annotated occlusion removal, occlusion removal in a specific context, and image adjustment for producing visually-pleasing images.

Following the proposal of GANs \cite{goodfellow2014generative}, extensive research has been carried out in GAN-based image modeling. GAN-based inpainting is one such application, where output images are conditioned on occluded input images \cite{pathak2016context, iizuka2017globally}. Conventional approaches for image inpainting, which do not use deep network based learning approaches \cite{ballester2001filling, efros2001image}, perform reasonably well for homogeneous image patches, but they fail in natural images. Yu \emph{et al.} \cite{Yu_2018_CVPR} recently proposed contextual attention in dilated CNN, which performs well for non-homogeneous patches. Moreover, several approaches for inpainting in a specific context, for instance, either face \cite{li2017generative}, head \cite{sun2018natural} or eye \cite{Dolhansky_2018_CVPR}, and for occlusion removal in a specific context, for instance, rain or shadow \cite{Liu_2018_CVPR, wang2018stacked}, have been proposed recently. Existing inpainting and occlusion removal techniques provide visually-pleasing results, but require manually annotating occluded regions and are not generic.

Recently proposed SEGAN \cite{ehsani2018segan} is partially aligned with our work. It relies on a semantic segmentator and a GAN to regenerate obscure content of objects in images, completing their appearance. Shotton \emph{et al.} \cite{shotton2009textonboost} present a novel approach for automatic semantic segmentation and visual understanding of images, which involves learning a discriminative model of object classes, interactive semantic segmentation and interactive image editing. Other related work include object recognition in cluttered environments \cite{jiang2016novel} and occlusion-aware layer-wise modeling of images \cite{huang2015efficient}.

These methods of enhancing occluded images either reconstruct images based on manually-annotated occluded regions or remove a specific type of occlusions in a particular context. However, none of these approaches declare occlusions based on image context. In other words, these do not address occlusion removal as a context-aware generic problem. In contrast, our approach involves making intelligent decisions on occlusions: detecting objects as occlusions depending on the image context, characterized by foreground and background objects, and regenerating occluded patches coherently. Thus, what we propose is an architecture for context-aware automatic occlusion removal in a generic domain, based on a fusion of image and language processing.\vspace{-4mm}

%% file: systemarchitecture.tex
\vspace{-2mm}
\begin{figure}[t]
	\centering
	\includegraphics[width=0.47\textwidth]{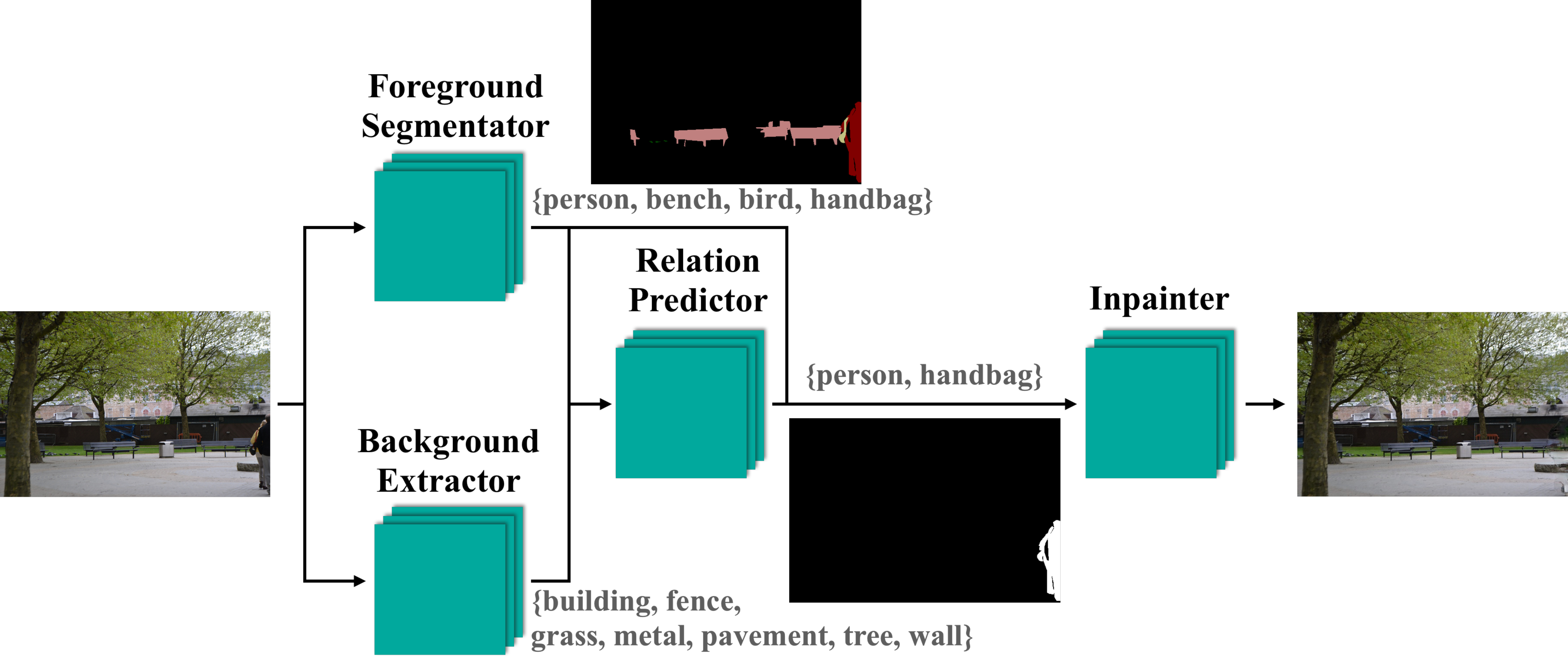}\vspace{-2mm}
	\caption{System Architecture}
	\label{Fig:sys_archi}\vspace{-6mm}
\end{figure}
\vspace{-2mm}Our system consists of four interconnected sub-networks as in Fig. \ref{Fig:sys_archi}: a foreground segmentator, a background extractor, a relation predictor, and an inpainter. An input image is fed into both the foreground segmentator and the background extractor. The foreground segmentator outputs pixel-wise association of foreground objects, or what we refer to as \textit{thing} classes, whereas the background extractor predicts background objects present in the image, commonly referred to as \textit{stuff} classes. These two components are followed by the relation predictor which utilizes both previously extracted thing and stuff classes.  Relation of each thing class to the image context is evaluated, and unrelated thing classes, i.e., occlusions, are provided as the output of the relation predictor. Finally, the image inpainter network exploits the relation predictions together with the pixel associations of the thing classes, to mask and regenerate pixels which belong to the occluding thing classes, generating an occlusion-free image. Following sub-sections elaborate more on the sub-networks in our proposed system. \vspace{-5mm}

\subsection{Sub-networks}\vspace{-2mm}
\label{ss:net_components}

Our foreground segmentator is a semantic segmentation network based on DeepLabv3+ \cite{chen2018encoder}, which achieves state-of-the-art performance on the PASCAL VOC 2012 semantic image segmentation dataset \cite{everingham2010pascal} with an mIoU of $89\%$. Due to the highly optimized nature of this network, we do not consider suggesting further improvements. Instead, we train it on the COCO-Stuff dataset \cite{caesar2016coco}, which is more challenging due to the generality of context and the higher number of classes, compared to the VOC dataset.

The background extractor is a CNN based on ResNet-101  \cite{he2016deep} model which consists of 100 convolutional layers and a single fully-connected layer. Since the task of the background extractor is to predict background classes, i.e., stuff in an image, this can be modeled as a multi-task classification problem. Thus, when training this component, we optimize an error which considers the summation of class-wise negative log-likelihoods.

The relation predictor provides the intelligence on occlusions based on image context. The relation between objects is estimated based on vector embeddings of class labels trained with a language model. To this end, we adopt the model proposed by Mikolov \emph{et al.} \cite{mikolov2013distributed} to represent COCO labels as vectors. Training this model for our requirement is different to the conventional requirement, where the model is trained on a large corpus of text, essentially learning linguistic relationships and syntax. What we have is image captions: five per image from MS COCO \cite{lin2014microsoft}, and what we want is to learn relations between objects as they appear in images, not linguistic relationships as in conventional models. Thus, we initially create a corpus concatenating image captions as the training set of the dataset. Importantly, the model should not learn the relations between any objects in separate images. Thus, we insert an end-of-paragraph (EOP) character at the end of each set of captions corresponding to a single image. Depending on the window size used in the model, the number of EOP characters inserted in between two sets of captions changes. Moreover, since we require the relation between object classes, not in a linguistic sense, it is logical to modify the corpus, removing all words except object classes. This allows the model to learn stronger relations between classes. Therefore, we train a separate model on this modified corpus. To visualize the embedding vectors of 128 dimensions in a 2D space, we use t-SNE algorithm \cite{maaten2008visualizing}. We use these word embeddings generated by the word-to-vector model trained on the modified captions of COCO-Stuff, to predict the relation between objects. What we want to quantify is ``how-related'' each thing class in an image to its context. We objectively capture this measure based on both thing and stuff classes, which can be represented as \vspace{-2mm}
\begin{equation}
\nonumber
d_{i\in\mathcal{T}_k}=\frac{1}{|\mathcal{S}_k\cup\mathcal{T}_k\backslash\{i\}|}\sum_{j\in\mathcal{S}_k\cup\mathcal{T}_k\backslash\{i\}}\cos_{\textnormal{sim}}\left(v_i\;,\;v_j\right),
\vspace{-2mm}
\end{equation}

\noindent where $\mathcal{T}_k$ and $\mathcal{S}_k$ represent the sets of thing and stuff classes of the image $k$. Here, $v$ represents an embedding vector of a class and $\cos_{\textnormal{sim}}\left(.,.\right)$, the cosine similarity between two vectors. The intuition here is to capture a similarity score between each embedding vector of a thing class and the remaining classes in an image, both thing and stuff. 

We base our inpainter on a recently proposed generative image inpainter \cite{Yu_2018_CVPR}, which is attentive to contextual information of images. This follows a two-stage architecture: coarse to fine. The first stage coarsely fills the mask, which is trained with a spatially discounted reconstruction $l_1$ loss. The second refines the generated pixels, which is trained with additional local and global WGAN-GP \cite{iizuka2017globally} adversarial loss. Contextual attention is imposed into the generation of missing pixels by implementing a special fully-convolutional encoder, which compares the similarity of patches around each generated pixel and patches in unmasked region to draw contextual information from the region having highest weighted similarity. Instead of training the inpainter with simulated square masks, we consider generating masks of random shapes and sizes utilizing the segmentation ground truth masks of thing classes. This provides better convergence for the network at the task of removing thing classes based on image context.\vspace{-4mm}

\subsection{Implementation Details}\vspace{-2mm}
\label{ss:implement}

\begin{figure*}[t]
	\centering
	\includegraphics[width=0.8\textwidth]{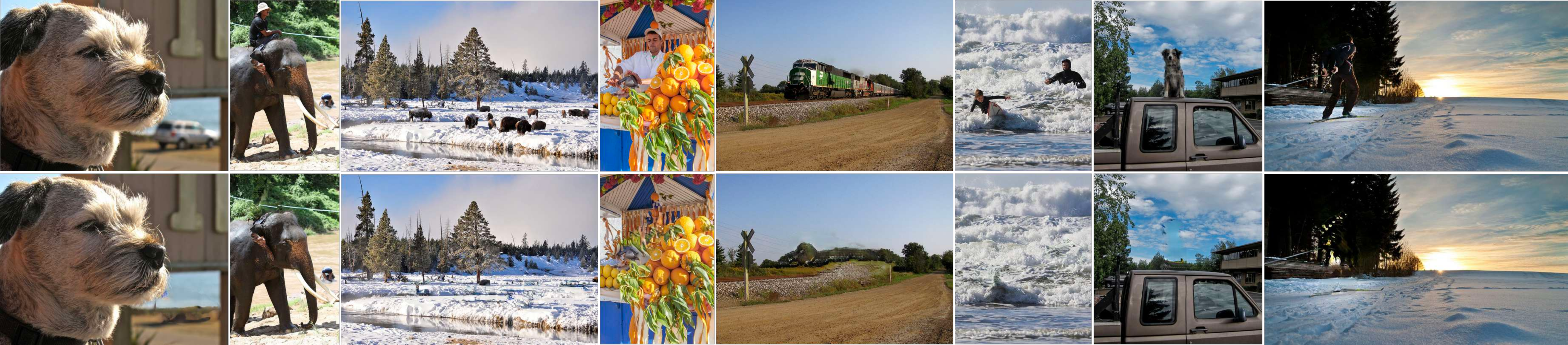}\vspace{-2mm}
	\caption{Some qualitative results of our system. A variety of thing classes has been detected as occlusions based on the context. For instance, note the two sets of pictures with a dog and a vehicle as thing classes. The vehicle is detected as an occlusion in one, while the dog is detected in the other based on the context.}
	\label{Fig:good_results}\vspace{-7mm}
\end{figure*}
Our system depends on image context to detect and remove occlusions. Thus, the dataset that we train our system on, should contain rich information about image context, which is why we choose COCO-Stuff dataset. In addition to the annotations of 91 foreground classes in MS COCO, this contains annotations of 91 background classes called stuff. This dataset includes 118K training samples and 5K validation samples. The large number of annotated images over different contexts enables better training of the system to extract contextual information. We utilize the semantic segmentations to train the foreground segmentator and the inpainter, the stuff class labels to train the background extractor, and the image captions to train the relation predictor.

In the foreground segmentator, we use a DeepLabv3+ model pre-trained on Pascal VOC, without initializing the last layer to accommodate the change in the number of classes. 
we train the segmentator for 125K iterations on COCO-Stuff. The background extractor based on ResNet-101 is trained from scratch for 50 epochs with multi-class classification loss and an input resoution of $128\times128$. The inpainter network is trained from scratch for 1000 epochs with random masks. For the relation predictor, we use the skip-gram model with a window size of 3  and train for 100K iterations on the corpus of image captions to learn 128-dimensional word embeddings.

To compensate for the aggregation of errors, we implement a few measures when interconnecting the networks\footnote{Source and trained models are available \href{https://github.com/kumarak93/occlusion_removal}{here} on GitHub.}. For instance, we handle inconsistent masks by dilation, and false-positives which are small in area by discarding the masks less than $2\%$ of the image. When considering the output of the relation predictor to identify occlusions, we consider a similarity threshold, which is set to be $0.4$. We compare the cosine similarity score of each thing class, normalized to a range of 0 to 1, and declare an occlusion when the similarity is less than this threshold.
\vspace{-4mm}

%% file: evaluation.tex
\vspace{-4mm}Our system automatically detects and removes unrelated objects, i.e., occlusions in input images. Thus, to evaluate the performance of the system, we raise two questions: which objects are removed by the system, and how good are the reconstructed images. Answering these evaluates the system for its utility in the application scenario. However, the evaluation of the system as a whole is challenging. This is because there is no publicly available dataset which annotates objects not related to an image context, or which contains images both with and without such occlusions. Thus, we choose to evaluate the system intuitively as presented below.\vspace{-4mm}

\subsection{Effectiveness of  Word-Embeddings}\vspace{-2mm}

\begin{figure}[t]
	\centering
	\includegraphics[width=0.42\textwidth]{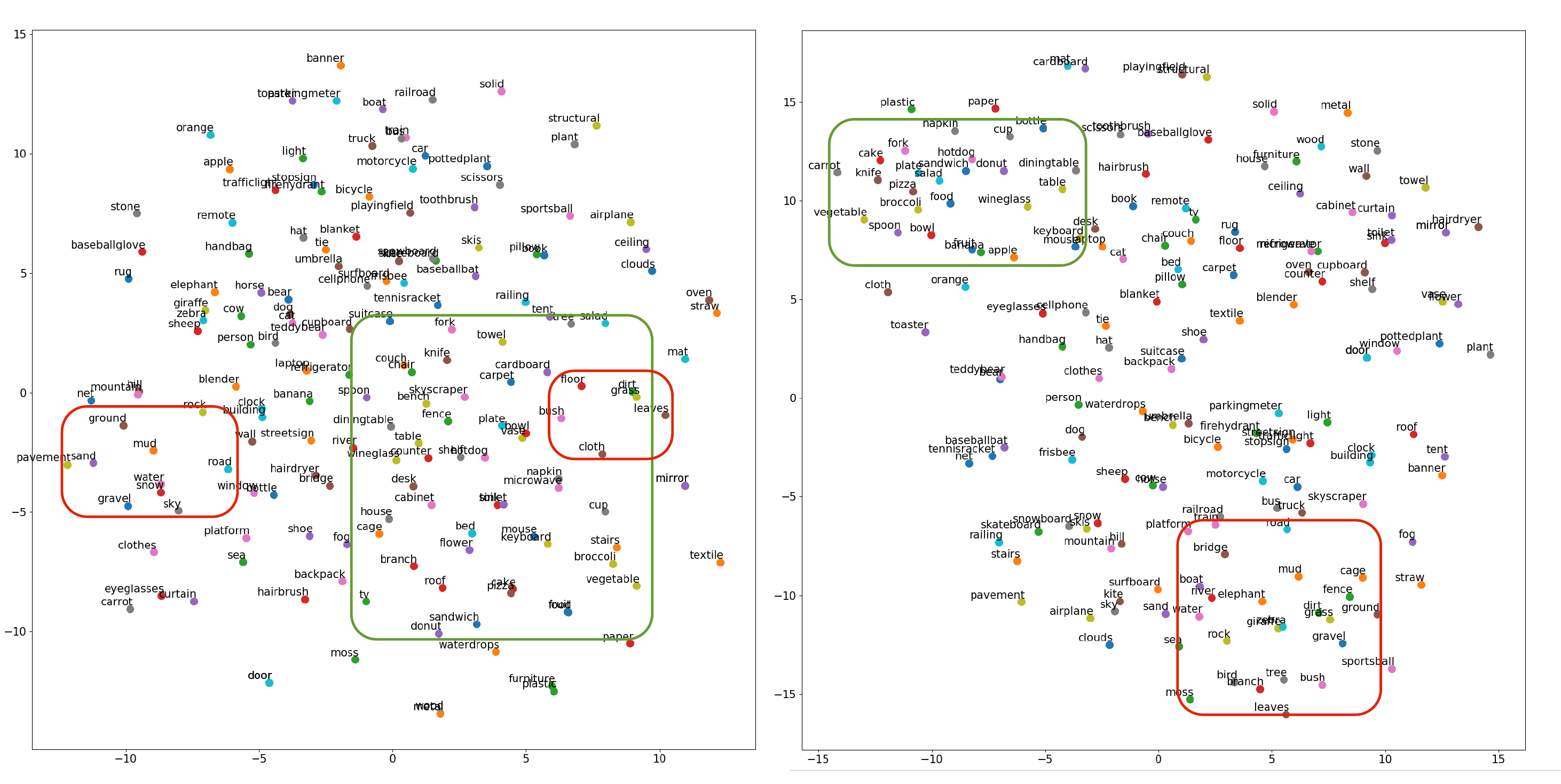}\vspace{-3mm}
	\caption{Embedding vectors mapped into 2D space. Left: trained on the original corpus, Right: trained on the modified corpus. The green and red clusters contain some related indoor objects and outdoor objects respectively, for which, the version on right shows better distinction.}
	\label{Fig:tsne}\vspace{-6mm}
\end{figure}

The relation predictor is responsible for making intelligent decisions on occlusions in images based on word-embeddings. Thus, it is useful to evaluate the effectiveness of the trained embedding-vectors. To qualitatively evaluate the embeddings, we map the 128-dimensional vectors to 2-dimensional space using t-SNE to visualize in a graph. Fig. \ref{Fig:tsne} shows the mapped embedding vectors in a 2D plane, separately trained on the original corpus of image captions and the modified corpus. The embeddings on the right show more meaningful mapping: having strongly related objects being mapped closer. For instance, it shows more distinction between some related indoor and outdoor object clusters. This is because the modified corpus contains only the thing and stuff class labels, enabling a stronger relation learning between classes.

We consider the relation between each pair of classes calculated as a cosine similarity, to quantitatively evaluate the effectiveness of the embeddings, comparing against a set of ground truth values. However, such ground truth data is not available and, generating such in wild would not be unbiased due to its subjectiveness. Thus, we intuitively consider a measure of relation using the count of occurrences of each object in dataset, which can be represented as\vspace{-2.5mm}
\begin{equation}
\nonumber
\mathcal{R}_{ab}=\frac{n_{a\cap b}}{n_{a\cup b}},\vspace{-2.5mm}
\end{equation}
where $a\cap b$ denotes the set of images which consists both $a$ and $b$ classes, and  $a\cup b$, the set of images which consists either of the classes. To capture the similarity between the relations from learned embeddings and counts, we evaluate their Pearson correlation, which yields a value of 0.527. This shows that our approach of estimating the relation has captured the co-existence of classes up to a certain extent. \vspace{-5mm}

\subsection{User Study}\vspace{-2mm} 
We conduct two user studies to evaluate the performance of the complete system, involving 1245 image pairs: original and occlusion-free reconstructed images in the validation set of COCO-Stuff. The two studies evaluate the visual-pleasing nature of the reconstructed images and a comparison between the occlusions identified by the users and the system. The former study shows users only the occlusion-free images and allows them to choose between the options: visually-pleasing or not, whereas the latter shows users only original images together with the thing class labels and allows them to record their opinion on the unrelated objects in the image context. Results of these user studies are presented in Table \ref{Tab:user}.

\begin{table}[t]
	\begin{center}
		\begin{tabular}{p{1.5cm}|p{1.5cm} p{1mm} p{1.5cm}|p{1.5cm}}
			\cline{1-2}\cline{4-5}
			\multicolumn{2}{c}{Visually-pleasing}&&\multicolumn{2}{c}{Relation}\\
			\cline{1-2}\cline{4-5}
			 Positive &  $992/1245$ $\mathbf{79.7\%}$&&Precision & $39.03\%$\\
			  Negative & $253/1245$  $\mathbf{20.3\%}$&&Recall & $17.46\%$\\
			\cline{1-2}\cline{4-5}
		\end{tabular}
		\caption{Preferences from the user study\vspace{-3mm}}
		\label{Tab:user}\vspace{-10mm}
	\end{center}
\end{table}
The majority of our reconstructed images has been rated visually-pleasing in the user study. This shows the effectiveness of the foreground segmentator and the inpainter, in accurately predicting the masks and reconstructing with attention to contextual details.
Although the precision and recall of the objects that our system detected as occlusions, in comparison with the user preferences is not high, it establishes a baseline as the first of its kind. This shows the deviation of our learning algorithm form actual human perception in some cases. In other words, in our proposed method, we consider learning relations from scratch, based only on image captions without any human annotations on relation.  In contrast, humans inherit the intuition of natural relation that comes from experience, which can be different to what is learned by the system. However, as shown in Fig. \ref{Fig:good_results}, a diverse set of objects has been detected as occlusions based on different image contexts. \vspace{-4mm}

%% file: conclusion.tex
\vspace{-4mm}We proposed a novel methodology for automatic detection and removal of occluding unrelated objects based on image context. To this end, we utilize vector embeddings of foreground and background objects, trained on modified image captions, to capture the image context by predicting the relation between objects. Although our approach learns meaningful relations between object classes and utilizes a hand-designed algorithm to decide on occlusions, human perception of it can be different. However, we establish a baseline for context-aware automatic occlusion removal in a generic domain, even without a publicly available dataset on relation. As future work, we hope to develop a dataset that captures human annotations on object relations, which will enable end-to-end training of such networks, improving our baseline on context-aware occlusion removal in images.\\ \vspace{-4mm}

\textbf{Acknowledgments:} K. Kahatapitiya was supported by the University of Moratuwa Senate Research Committee Grant no. SRC/LT/2016/04 and D. Tissera, by QBITS Lab, University of Moratuwa. The authors thank  the Faculty of Information Technology of the University of Moratuwa, Sri Lanka for providing computational resources.